\documentclass[11pt]{article}
\usepackage[preprint]{acl}
\usepackage{times}
\usepackage{latexsym}
\usepackage[T1]{fontenc}
\usepackage[utf8]{inputenc}
\usepackage{microtype}
\usepackage{graphicx}
\usepackage{xurl}
\usepackage{placeins}
\usepackage{tikz}
\usepackage[nosfdefault]{comicneue}
\usetikzlibrary{arrows.meta}
\usepackage{booktabs}
\usepackage{amsmath,amssymb}
\usepackage{xcolor}
\newcommand{\frank}[1]{{#1}}
\title{Training Negatives Still Matter After Calibrating Answerability Readouts}
\author{
  Jinchen Zhang \quad Zheyuan Liu \quad Yihan Zhu \quad Meng Jiang \\
  University of Notre Dame \\
  \texttt{jzhang47@nd.edu}
}
\hypersetup{
  pdftitle={Training Negatives Still Matter After Calibrating Answerability Readouts},
  pdfauthor={Jinchen Zhang, Zheyuan Liu, Yihan Zhu, Meng Jiang}
}

\begin{document}
\maketitle
\begin{abstract}
Answerability readouts predict whether a passage supplies an answer, but whether calibration can compensate for removing negative passages during training remains unclear. We present a controlled study of how training-negative construction affects calibrated prediction, ranking, and downstream answer retention. We compare native training, which retains negative passages, with empty training, which removes them, holding questions, labels, positive examples, and model weights fixed. Both are calibrated on the same labeled native examples. On SQuAD 2.0 and RepLiQA, native-trained logistic readouts at selected depths achieve lower Brier loss for Qwen2.5-7B-Instruct and Gemma-2-9B-it, with or without output scores. Native fusion also outperforms calibrated output alone. Post-hoc AUROC analysis shows better ranking under native training, beyond what threshold adjustment alone can explain. On the official SQuAD 2.0 development set, native readout policies improve F1 over empty policies but retain more answers to unanswerable questions. The advantage varies by estimator and depth, while RepLiQA's text shortcuts limit claims about contextual understanding. Unfinished generations prevent complete answer-level evaluation, although worst-case bounds preserve the native–empty F1 gains. Our findings show that training negatives affect ranking and downstream decisions even after common calibration, without establishing a universal advantage for native training.
\end{abstract}

\section{Introduction}

\begin{figure}[!t]
\centering
\definecolor{ovink}{HTML}{334155}
\definecolor{ovline}{HTML}{AAB4BF}
\definecolor{ovwhite}{HTML}{FFFFFF}
\definecolor{ovblue}{HTML}{2171B5}
\definecolor{ovbluefill}{HTML}{F0F6FB}
\definecolor{ovbluefold}{HTML}{D9E9F6}
\definecolor{ovorange}{HTML}{B86C2C}
\definecolor{ovorangefill}{HTML}{FCF5EE}
\definecolor{ovorangefold}{HTML}{F4E0CF}
\definecolor{ovteal}{HTML}{347C78}
\definecolor{ovtealfill}{HTML}{F0F7F5}
\definecolor{ovgrayfill}{HTML}{F6F8FA}
\definecolor{ovskin}{HTML}{F4C892}
\definecolor{ovhair}{HTML}{59616B}
\definecolor{ovrobot}{HTML}{FFF2C6}
\begin{tikzpicture}[x=1mm,y=-1mm,line width=.65pt,
  font=\comicneue,text=ovink,line cap=round,line join=round]
\path[use as bounding box] (0,0) rectangle (76,85);
\path[draw=ovink,fill=ovwhite,rounded corners=1.8mm] (0.2,0.2) rectangle +(75.6,19.6);
\path[draw=ovink,fill=ovink] (3.8,3.7) circle[radius=2.15mm];
\node[font=\comicneue\fontsize{9}{9.9}\selectfont\bfseries,text=ovwhite,inner sep=0pt,align=center] at (3.8,3.6999999999999997) {1};
\node[font=\comicneue\fontsize{10}{11.0}\selectfont\bfseries,text=ovink,inner sep=0pt,align=left,anchor=west] at (7.8,3.7) {One question};
\path[draw=ovink,fill=ovteal] (3.7,18.1)--(4,15.5)--(5.4,14.4)--(7.5,14.4)--(8.9,15.5)--(9.2,18.1)--cycle;
\path[draw=ovink,fill=ovskin] (6.5,11.5) circle[radius=2.3mm];
\path[draw=ovhair,fill=ovhair] (4.3,11.2)--(4.4,9.8)--(5.6,8.9)--(7.5,9.2)--(8.6,10.3)--(7.3,10.4)--(5.7,9.9)--cycle;
\path[draw=ovink,fill=ovink] (5.7,11.7) circle[radius=0.14mm];
\path[draw=ovink,fill=ovink] (7.3,11.7) circle[radius=0.14mm];
\path[draw=ovink] (5.8,12.7)--(6.5,13)--(7.2,12.7);
\path[draw=ovline,fill=ovgrayfill,rounded corners=1.5mm] (13,7) rectangle +(60,11.4);
\path[draw=ovline,fill=ovgrayfill] (13,11.4)--(10.4,12.7)--(13,14);
\path[draw=ovtealfill,fill=ovtealfill,rounded corners=0.6mm] (31,13.1) rectangle +(24,4);
\node[font=\comicneue\fontsize{9}{9.9}\selectfont,text=ovink,inner sep=0pt,align=center] at (43.0,12.7) {Who aligned the country with the\\\textbf{Western Bloc}?};
\path[draw=ovink,fill=ovwhite,rounded corners=1.8mm] (0.2,22.2) rectangle +(75.6,40.6);
\path[draw=ovink,fill=ovink] (3.8,25.7) circle[radius=2.15mm];
\node[font=\comicneue\fontsize{9}{9.9}\selectfont\bfseries,text=ovwhite,inner sep=0pt,align=center] at (3.8,25.7) {2};
\node[font=\comicneue\fontsize{10}{11.0}\selectfont\bfseries,text=ovink,inner sep=0pt,align=left,anchor=west] at (7.8,25.7) {Two training inputs};
\path[draw=ovblue,fill=ovbluefill,rounded corners=1.3mm] (3,29) rectangle +(33,27);
\node[font=\comicneue\fontsize{9.5}{10.450000000000001}\selectfont\bfseries,text=ovblue,inner sep=0pt,align=center] at (19.5,32.1) {Native};
\path[draw=ovblue,fill=ovwhite] (5,35)--(32.4,35)--(34,36.6)--(34,48.2)--(5,48.2)--cycle;
\path[draw=ovblue,fill=ovbluefold] (32.4,35)--(32.4,36.6)--(34,36.6)--cycle;
\path[draw=ovorange,fill=ovorangefill,dash pattern=on 1.3pt off 1.0pt,rounded corners=1.3mm] (40,29) rectangle +(33,27);
\node[font=\comicneue\fontsize{9.5}{10.450000000000001}\selectfont\bfseries,text=ovorange,inner sep=0pt,align=center] at (56.5,32.1) {Empty};
\path[draw=ovorange,fill=ovwhite,dash pattern=on 1.3pt off 1.0pt] (42,35)--(69.4,35)--(71,36.6)--(71,48.2)--(42,48.2)--cycle;
\path[draw=ovorange,fill=ovorangefold] (69.4,35)--(69.4,36.6)--(71,36.6)--cycle;
\node[font=\comicneue\fontsize{8.7}{9.57}\selectfont,text=ovink,inner sep=0pt,align=center] at (19.5,37.35) {Sassou Nguesso};
\node[font=\comicneue\fontsize{8.5}{9.350000000000001}\selectfont,text=ovink,inner sep=0pt,align=center] at (19.5,40.6) {aligned with};
\path[draw=ovbluefold,fill=ovbluefold,rounded corners=0.6mm] (7.5,42.7) rectangle +(24,4.2);
\node[font=\comicneue\fontsize{9}{9.9}\selectfont\bfseries,text=ovblue,inner sep=0pt,align=center] at (19.5,44.800000000000004) {Eastern Bloc};
\node[font=\comicneue\fontsize{8.2}{9.02}\selectfont,text=ovink,inner sep=0pt,align=center] at (19.5,52.2) {Passage present,\\answer unsupported};
\node[font=\comicneue\fontsize{9}{9.9}\selectfont,text=ovorange,inner sep=0pt,align=center] at (56.5,41.6) {No passage};
\path[draw=ovorange,fill=ovwhite] (66.2,32.7) circle[radius=0.7mm];
\path[draw=ovorange,fill=ovwhite] (67.4,33.7) circle[radius=0.7mm];
\path[draw=ovorange] (66.5,32.2)--(71.2,29.9);
\path[draw=ovorange] (67.8,33.2)--(69.6,29.3);
\path[draw=ovorange,fill=ovorange] (68.3,31.6) circle[radius=0.22mm];
\node[font=\comicneue\fontsize{8.2}{9.02}\selectfont,text=ovink,inner sep=0pt,align=center] at (56.5,52.2) {Passage removed,\\label unchanged};
\path[draw=ovgrayfill,fill=ovgrayfill,rounded corners=1mm] (3,57.3) rectangle +(70,4.5);
\path[draw=ovink,fill=ovwhite] (5.1,58.55)--(8.7,58.55)--(10,59.55)--(8.7,60.55)--(5.1,60.55)--cycle;
\path[draw=ovink,fill=ovink] (8.7,59.55) circle[radius=0.2mm];
\node[font=\comicneue\fontsize{8.9}{9.790000000000001}\selectfont\bfseries,text=ovink,inner sep=0pt,align=center] at (40.5,59.55) {Both labels: unanswerable};
\path[draw=ovink,fill=ovwhite,rounded corners=1.8mm] (0.2,65.2) rectangle +(75.6,19.6);
\path[draw=ovink,fill=ovink] (3.8,68.7) circle[radius=2.15mm];
\node[font=\comicneue\fontsize{9}{9.9}\selectfont\bfseries,text=ovwhite,inner sep=0pt,align=center] at (3.8,68.7) {3};
\node[font=\comicneue\fontsize{10}{11.0}\selectfont\bfseries,text=ovink,inner sep=0pt,align=left,anchor=west] at (7.8,68.7) {At evaluation};
\path[draw=ovteal,fill=ovwhite] (6.3,72)--(9.819999999999999,72)--(10.7,72.88)--(10.7,76.8)--(6.3,76.8)--cycle;
\path[draw=ovteal,fill=ovtealfill] (9.819999999999999,72)--(9.819999999999999,72.88)--(10.7,72.88)--cycle;
\path[draw=ovteal] (7.1,73.8)--(9.9,73.8);
\path[draw=ovteal] (7.1,75)--(9.9,75);
\node[font=\comicneue\fontsize{8.6}{9.46}\selectfont,text=ovteal,inner sep=0pt,align=left,anchor=west] at (15.5,74.4) {New questions with passages};
\path[draw=ovline,fill=ovgrayfill] (8.5,80.7) circle[radius=1.9mm];
\node[font=\comicneue\fontsize{8.5}{9.350000000000001}\selectfont\bfseries,text=ovink,inner sep=0pt,align=center] at (8.5,80.7) {?};
\node[font=\comicneue\fontsize{8.9}{9.790000000000001}\selectfont,text=ovink,inner sep=0pt,align=left,anchor=west] at (15.5,80.69999999999999) {Can common calibration compensate\\for empty-negative training?};
\end{tikzpicture}
\caption{A SQuAD fitting example with the passage summarized.
Native (solid blue) and empty (dashed orange) training share the same
unanswerable question and label. The bottom panel concerns calibration on
new inputs, not predictions for this example.}
\label{fig:overview}
\end{figure}
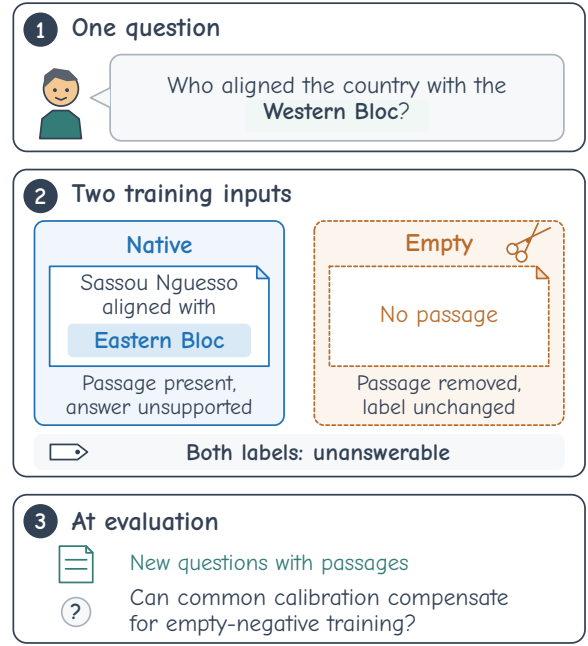

A passage can discuss the subject of a question without providing its
answer. A question-answering system instructed to rely on that passage
must distinguish the two cases to decide when to abstain. An
\emph{answerability readout} predicts whether the supplied text supports
an answer, providing a signal for that decision
\citep{slobodkin2023answerability,lavi2026directions}.

\frank{For hidden-state answerability readouts, threshold
adjustment can improve cross-dataset transfer without changing the learned direction}
\citep{lavi2026directions}.
\frank{Work on constructed negatives also examines transfer
with label-free thresholds} \citep{luo2026yopo}.
\frank{A simple way to construct a negative example is to
delete the passage from an unanswerable question--passage pair, but the
readout must still recognize unsupported questions when text is present
at evaluation.
Prior results on threshold transfer do not establish whether
calibration on the same labeled question--passage examples can compensate
for omitting that context during fitting.}

\begin{figure*}[t]
\centering
\input{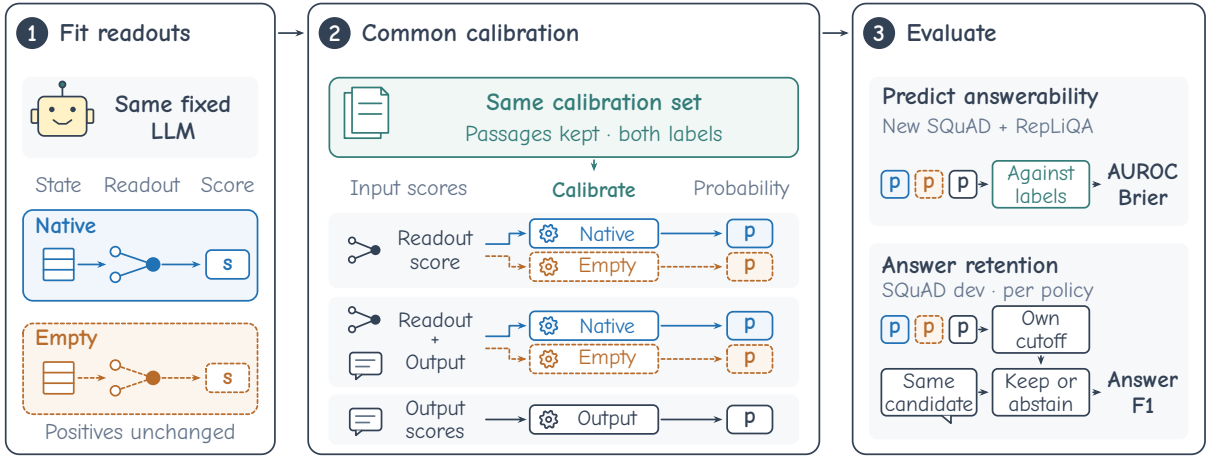}
\caption{Controlled comparison of native (solid blue) and empty
(dashed orange) fitting, with the model, readout depth, questions, and
labels held fixed. The left panel fits readouts on condition-specific
states, yielding scores $s$.
Neural and speech icons denote readout and output scores, respectively.
Separate maps (gears) use identical labeled native examples (teal) to
calibrate readout scores, both signals jointly (fusion), or output scores
alone (gray) into probabilities $p$. Prediction quality uses new SQuAD
articles and RepLiQA documents. Answer retention uses official SQuAD 2.0
development questions. Each probability-based veto applies its own threshold to $p$ to
keep the same generated candidate within a model or abstain.
Primary Brier comparisons use fusion, and primary
answer-retention comparisons use readouts alone.}
\label{fig:design}
\end{figure*}

\frank{To address this question, we compare \emph{native negatives}, which retain the original
non-answering passages, with \emph{empty negatives}, which remove them
(Figure~\ref{fig:overview}).
We hold questions, labels, positive examples, and model
weights fixed to isolate the effect of retaining negative passages.
Both conditions share depth selection on native inputs
and use identical labeled native examples to fit separate calibration
maps (Figure~\ref{fig:design}). We evaluate Qwen2.5-7B-Instruct and
Gemma-2-9B-it on SQuAD 2.0 and RepLiQA.
Our contributions are summarized as follows:}

\begin{itemize}
\item \textbf{Training negatives affect calibrated
prediction.}
At the selected depths, native-trained logistic readouts have lower
Brier loss than empty-trained readouts on both datasets. The Brier
advantage also holds for \emph{fusion}, which combines each readout with an output score
derived from the likelihood of the prescribed abstention response.
Adding the native readout to the output score also lowers Brier loss
relative to calibrated output alone (\S\ref{sec:qa-predictions}).

\item \textbf{The predictive advantage includes better
input ranking.}
Post-hoc comparisons of the calibrated readouts without output scores
show higher AUROC under native training in both models and datasets
(\S\ref{sec:ranking}). Threshold adjustment alone cannot remove this
ranking difference, so the advantage is not solely a consequence of
threshold placement.

\item \textbf{Native fitting changes the trade-off between
answering and abstaining.}
On the official SQuAD 2.0 development set, readout-based policies
retain each model's fixed candidate or abstain.
Native policies achieve higher F1 than empty policies in both models,
retaining more correct answers but also more answers to unanswerable
questions. Worst-case bounds for unfinished generations keep
the native--empty F1 gains positive. Native gains over unchanged behavior (Qwen) and
output-based vetoes (Gemma) remain inconclusive
(\S\ref{sec:answers}).
\end{itemize}

The predictive advantage of native fitting depends on
estimator and depth, with some settings yielding lower fusion Brier loss
under empty fitting. On RepLiQA, a word-count classifier also
benefits from native training, so the construction effect is not specific
to hidden-state readouts
(\S\ref{sec:boundaries}).

\section{\frank{Motivation}}
\label{sec:motivation}
\frank{In Figure~\ref{fig:overview}, the negative example
pairs a passage about the Eastern Bloc with a question about the Western
Bloc. Native fitting retains this topically related passage, whereas
empty fitting removes it while keeping the question and negative label.
Deletion changes content,
length, and text presence together, so the comparison cannot attribute
an advantage specifically to learning semantic mismatch.}

\frank{Removing negative passages may affect the threshold
needed for answerability decisions, the ranking of inputs, or both.
We compare probability quality after common probability calibration to
test whether a predictive difference remains. Ranking provides a
complementary test because threshold adjustment cannot change the
ordering of inputs. Because passage support does not guarantee a
correct generated answer, we also examine answer retention.}

\section{\frank{Related Work}}
\paragraph{\frank{Answerability readouts.}}
\frank{Hidden-state classifiers predict answerability
\citep{slobodkin2023answerability}, and sparse-autoencoder studies examine
feature generalization for this task
\citep{heindrich2025answerability}.
\citet{lavi2026directions} compare difference-in-means (DiM) and logistic readouts on four QA
datasets, including SQuAD 2.0 and RepLiQA, and find that validation-set threshold adjustment improves
cross-dataset transfer without changing the learned direction.
\citet{luo2026yopo} study native and constructed negatives, including
document replacement and answer-sentence deletion, with label-free
quantile transfer thresholds.
We vary negative fitting passages and evaluate ranking and prediction
after calibration on identical labeled native examples.}

\paragraph{\frank{Negative construction and shortcuts.}}
\frank{For tested QA systems, SQuAD's human-written negatives are harder than
TF-IDF-retrieved or rule-generated negatives in separate datasets
\citep{rajpurkar2018squad2}.
BM25 hard negatives improve in-batch dense passage retrieval
\citep{karpukhin-etal-2020-dense}.
Partial-input baselines expose shortcuts in reading comprehension
\citep{kaushik-lipton-2018-much} and NLI
\citep{gururangan-etal-2018-annotation,poliak-etal-2018-hypothesis}.
Our question-only and word-count controls test label shortcuts.}

\paragraph{\frank{Irrelevant context.}}
\frank{Irrelevant text can disrupt comprehension \citep{jia-liang-2017-adversarial}
and reasoning \citep{shi2023irrelevant}, motivating context-robust training
\citep{yoran2024irrelevant}.
\citet{zhang-etal-2026-confidence-estimation} test whether multi-turn
confidence responds to useful evidence or conversational filler.
We compare ranking and threshold transfer with evidence and token-matched
irrelevant additions.}

\paragraph{\frank{Calibration and abstention.}}
\frank{Calibration has been studied for neural classifiers
\citep{guo2017calibration} and language-model QA \citep{jiang-etal-2021-know}.
Confidence methods use self-evaluation or variation across generations
\citep{kadavath2022know,farquhar2024entropy,geng-etal-2024-survey}.
Selective classification relates rejection to
prediction risk and coverage \citep{geifman2017selective,franc2023reject}.
\citet{kamath2020selective} predict answer correctness under domain shift,
whereas our veto signal predicts passage answerability.
\citet{kim-etal-2025-speak} change decoding to answer or abstain according
to available knowledge. We test how negative-context fitting changes
answer retention with generated candidates fixed.}

\section{\frank{Background}}
\label{sec:background}
\noindent\frank{Each QA example $(q_i,c_i,y_i)$ contains a question, a passage,
and a dataset label $y_i\in\{0,1\}$. The label is 1 when the passage
supplies an answer. The formatted prompt $x_i$ requests a supported answer
or a fixed abstention response when the passage supplies no answer.
SQuAD requests the shortest copied span,
whereas RepLiQA allows a short paraphrase. Labels describe passage support
regardless of the model's generated answer. Appendix~\ref{app:qa-prompts}
gives both templates.}

\frank{Passage support and answer correctness are distinct targets:
sufficient context does not ensure successful model use
\citep{joren2025sufficient}, and answerability and correctness signals
can support different abstention decisions \citep{wagner2026axes}.
Probe research also distinguishes selectivity from accuracy
\citep{hewitt2019} and decodability from model use \citep{belinkov2022}.
We interpret readout performance as predictive access to passage
answerability, without inferring how the model uses that signal.}

\frank{Ranking, probability quality, and thresholded decisions measure
different properties of these predictions. Area under the receiver operating
characteristic curve (AUROC) measures the ranking of answerable and
unanswerable inputs. Probability calibration maps scores to probabilities using sigmoid
fitting \citep{platt2000probabilistic} or isotonic regression
\citep{zadrozny2002transforming}. A strictly increasing scalar map of final predictions
preserves AUROC, whereas isotonic regression can introduce
ties. To assess probability quality over $n$ evaluated inputs with final
answerability probabilities $p(x_i)\in[0,1]$, we use Brier loss,
$\mathrm{BS}=n^{-1}\sum_{i=1}^{n}(p(x_i)-y_i)^2$, a proper scoring rule
\citep{gneiting2007proper}. Balanced accuracy (BA) averages the two
class recalls after thresholding.}

\frank{A lower Brier loss can reflect less miscalibration or better
discrimination. Murphy's decomposition separates reliability, resolution,
and uncertainty \citep{murphy1973partition}. We use the CORP isotonic
decomposition \citep{dimitriadis2021corp}, whose miscalibration (MCB) and
discrimination (DSC) terms correspond in role to reliability and resolution,
respectively, without necessarily matching their numerical values.
Its evaluation-label fit is descriptive and is never deployed as a
correction (Appendix~\ref{app:decomposition}).}

\section{\frank{Negative-Context Comparison Protocol}}
\label{sec:comparison-protocol}
\frank{For each negative example, native fitting retains $c_i$ and empty fitting
leaves the passage field blank in the same prompt template.
Positive examples, questions, labels, and model weights remain fixed.
Depth selection, calibration, and evaluation retain both classes' original
passages (\emph{native inputs}, Figure~\ref{fig:design}).}

\paragraph{\frank{Readout scores.}}
\frank{Let $h_\ell(x)\in\mathbb{R}^{d_\ell}$ be the raw output of
zero-based block $\ell$ at the last prompt token, before generation,
where $d_\ell$ is the hidden-state dimension. For condition
$t\in\{\mathrm{native},\mathrm{empty}\}$, let $z_t(x)$ denote this state
standardized using condition $t$'s fitting-set coordinate statistics for
logistic regression. For DiM, $z_t(x)=h_\ell(x)$.
The QA readout score is
\begin{equation}
s_t(x)=w_t^\top z_t(x)+b_t,
\label{eq:readout}
\end{equation}
with $w_t\in\mathbb{R}^{d_\ell}$ and $b_t\in\mathbb{R}$.}
We fit L2-regularized logistic regression, a standard
answerability readout \citep{slobodkin2023answerability,lavi2026directions}.
\frank{DiM uses the normalized difference between positive and negative
raw activation means \citep{lavi2026directions}, with $b_t=0$ for QA.
Its scalar calibration map supplies the probability intercept.
Appendix~\ref{app:estimators} gives the fitting details.}

We select a shared depth by cross-validating logistic AUROC on native
inputs and averaging over both training conditions.
Among four candidates per model, we select zero-based
block 20 for Qwen and block 30 for Gemma on both datasets
(Appendix~\ref{app:depth-selection}). Validation against empty negatives
is nearly saturated across these depths in the SQuAD fitting data,
making it less informative for selection. DiM uses the same selected depth,
and final-block fits test depth dependence.

\paragraph{\frank{Common probability calibration.}}
\label{sec:readout-selection}
\frank{Each component pairs a fitted readout with a calibration map.
Separate scalar logistic maps $g_t:\mathbb{R}\to[0,1]$ calibrate
the two readouts on identical labeled native examples held out from
each component's weight fitting \citep{platt2000probabilistic}.
The \emph{output score} $u(x)$ sums negative conditional log probabilities of
the prescribed abstention response, without an end-of-sequence token.
Larger $u$ means less probability assigned to this response.
To test whether the output signal complements the readout,
a two-input logistic map $g_t^{\mathrm{fusion}}:\mathbb{R}^2\to[0,1]$
combines their scores. For one component, the probabilities are
\begin{equation}
\begin{aligned}
p_t(x)&=g_t\bigl(s_t(x)\bigr),\\
p_t^{\mathrm{fusion}}(x)&=g_t^{\mathrm{fusion}}\bigl(u(x),s_t(x)\bigr).
\end{aligned}
\label{eq:calibration}
\end{equation}
Calibration standardizes these scalar inputs using calibration data only.
The SQuAD predictor averages six calibrated component predictions,
whereas RepLiQA uses one fitted component per condition.}

\paragraph{\frank{Comparators.}}
For the output-only comparison, we use isotonic calibration on SQuAD
\citep{zadrozny2002transforming}. On RepLiQA, we select by calibration
Brier loss among \frank{maps that use either the score for the full
abstention response or separate scores for its tokens}
(Appendix~\ref{app:data-selection}). Answer-level evaluation instead
selects the output comparator by calibration answer F1
(\S\ref{sec:answer-protocol}).
To test whether simple text features predict the labels, we fit baselines
using question-only or question--passage TF-IDF features, or just the
numbers of words in the question and passage.

\section{\frank{Experiments}}
\label{sec:results}
\subsection{\frank{Experimental Setup}}
\label{sec:setup}
\frank{We use Qwen2.5-7B-Instruct and Gemma-2-9B-it
\citep{qwen2024,gemma2024} on SQuAD 2.0 \citep{rajpurkar2018squad2}
and RepLiQA \citep{monteiro2024repliqa}. RQ1 tests calibrated probability
quality, and RQ2 asks whether a ranking difference remains. RQ3 measures the
answering consequences on a separate split, and RQ4 examines estimator,
depth, and dataset boundaries. A constructed factual task provides
an additional threshold-transfer diagnostic.}

\paragraph{\frank{Data roles.}}
SQuAD extends span-based QA \citep{rajpurkar-etal-2016-squad} with
human-written unanswerable questions. We reserve its official development
split for answer-level evaluation. Probability evaluation uses 164 new
articles from the training release, totaling 1,312 class-balanced inputs.
\frank{The six SQuAD components use different assignments of a 48-article
pool. Each component separates weight fitting from probability calibration,
but components overlap and share depth selection on the full pool.
We average their probabilities on new articles without refitting.}
RepLiQA contains human-created fictional documents. We use its
\frank{first released split with 256 new evaluation documents and
1,280 questions, including 1,020 answerable and 260 unanswerable questions.}
Both conditions share fitting and calibration assignments within each study.
The studies use different fitting protocols and class proportions, so we
report them separately rather than pool their effects. We exclude evaluation
texts from fitting, selection, calibration, and earlier diagnostics.
Appendix~\ref{app:protocol} gives the assignments and study history.

\paragraph{\frank{Endpoints and uncertainty.}}
For each model, the two prespecified primary comparisons are native logistic fusion
against empty logistic fusion and against output alone. They test whether
fitting negatives matter when the output signal is available and whether
the readout adds value to that signal.
\frank{Brier loss is the primary endpoint. AUROC uses final probabilities,
including the SQuAD component average. For descriptive BA, we classify an
input as answerable when $p(x_i)>0.5$. Readout-alone AUROC differences have post-hoc
paired-cluster intervals (Appendix~\ref{app:auroc-uncertainty}).}
For the primary Brier comparisons, each article or document contributes
one mean paired loss difference.
We apply Bonferroni correction to the four comparisons per dataset, yielding
98.75\% $t$ intervals, and check sensitivity with paired cluster bootstrap
intervals. \frank{Both methods condition on fixed fits and calibration
maps, excluding refitting and selection uncertainty.}

\paragraph{\frank{Answer-level protocol.}}
\label{sec:answer-protocol}
Each model generates one greedy candidate, capped at 64 tokens, for all
11,873 questions in the official SQuAD 2.0 development set.
Every veto policy receives the same candidate and either retains it or
replaces it with abstention. None can recover an answer the model did
not generate.

Readout-based vetoes reuse the SQuAD readouts, calibration maps, and
component average to score each input from its pre-generation state.
\frank{Earlier limits on retaining answers to
unanswerable calibration questions also rejected many correct answers.
We therefore choose veto thresholds to maximize answer F1 on 128 generated
calibration answers per model, using articles held out from probability
calibration.
The objective and thresholds are fixed before development evaluation.}
To isolate the readout's answering consequences, the primary
policy uses the \emph{readout alone}, with fusion as a secondary comparison.
For each model, we compare the native readout policy with the empty
readout policy, an output-based veto, and unchanged model behavior.
For the output-based veto, calibration answer F1 selects a scalar
logistic map for Qwen and an isotonic map for Gemma.

We average official SQuAD F1, based on normalized token overlap, over all
development questions. We use question-weighted paired article-bootstrap
intervals adjusted for six comparisons.
With complete outputs, the prespecified criteria require a gain of at least
1 F1 point and an adjusted interval lower bound above zero.
\frank{Unfinished generations receive zero credit for every policy and remain
in the denominator. Sensitivity bounds quantify their effect, but the
prespecified complete-output requirement remains unmet
(Appendix~\ref{app:qa}).}

\subsection{\frank{RQ1: Does Native Fitting Improve Calibrated Prediction?}}
\label{sec:qa-predictions}
\begin{figure}[!b]
\centering
\includegraphics[width=\columnwidth]{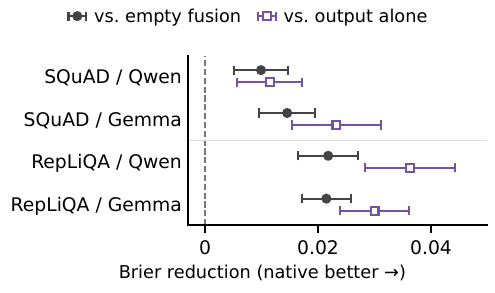}
\caption{Brier reductions from native fusion relative to empty fusion
(circles) and output alone (squares). Positive values favor native fusion.
Error bars show paired 98.75\% article/document $t$ intervals, corrected
for four comparisons per dataset. Datasets are not pooled.
\frank{Appendix~\ref{app:probability-results}}, Table~\ref{tab:confirmation}, gives exact differences and intervals.}
\label{fig:probability-effects}
\end{figure}

With calibration examples and output scores held fixed, native-trained
logistic readouts yield lower fusion Brier loss than empty-trained
readouts on both datasets (Table~\ref{tab:primary}).
On SQuAD, native rather than empty training reduces fusion Brier loss
from 0.095 to 0.085 for Qwen and from 0.110 to 0.095 for Gemma.
On RepLiQA the reductions are
0.022 and 0.021, respectively.
Native fusion also outperforms calibrated output alone. Paired $t$ and
paired cluster bootstrap intervals exclude zero for all four primary
comparisons per dataset (Figure~\ref{fig:probability-effects}).
RepLiQA's text shortcuts limit what those gains establish about relating
questions to passages (\S\ref{sec:boundaries}).

\begin{table}[!ht]
\centering\small
\begin{tabular}{@{}lrrr@{}}
\toprule
\multicolumn{1}{@{}l}{\begin{tabular}[c]{@{}l@{}}Dataset / model\end{tabular}} & \multicolumn{1}{c}{\begin{tabular}[c]{@{}c@{}}Native\\fusion\end{tabular}} & \multicolumn{1}{c}{\begin{tabular}[c]{@{}c@{}}Empty\\fusion\end{tabular}} & \multicolumn{1}{c@{}}{\begin{tabular}[c]{@{}c@{}}Output\end{tabular}} \\
\midrule
SQuAD / Qwen & 0.085 & 0.095 & 0.097 \\
SQuAD / Gemma & 0.095 & 0.110 & 0.119 \\
\midrule
RepLiQA / Qwen & 0.029 & 0.051 & 0.066 \\
RepLiQA / Gemma & 0.043 & 0.064 & 0.073 \\
\bottomrule
\end{tabular}

\caption{Primary Brier losses on new articles and documents
at selected depths. Lower is better. Fusion predictors share output
scores and labeled calibration examples, differing only in training
negatives. Output is each study's prespecified comparator.
Figure~\ref{fig:probability-effects} shows paired improvements.}
\label{tab:primary}
\end{table}

The advantage over output alone could reflect unequal access to
training labels or output features. In post-hoc SQuAD controls, we give
output calibration access to the readout-fitting articles and \frank{separate
token scores for the abstention response}. Native fusion still has lower Brier loss than all
six variants (Appendix~\ref{app:output-controls},
Table~\ref{tab:output-controls}), so expanded label and feature access
does not close the gap.

Native training lowers Brier loss even without the output score
(Table~\ref{tab:controls}), so the benefit does not require fusion.
On SQuAD, native training reduces readout-alone Brier from 0.106 to
0.088 for Qwen and from 0.121 to 0.099 for Gemma.
Fusion adds little to the native readouts, lowering SQuAD Brier by
at most 0.004 and slightly raising RepLiQA Brier.

\begin{table}[!ht]
\centering\small
\begin{tabular}{@{}llrrr@{}}
\toprule
\multicolumn{1}{@{}l}{\begin{tabular}[c]{@{}l@{}}Model\end{tabular}} & \multicolumn{1}{l}{\begin{tabular}[c]{@{}l@{}}Training\end{tabular}} & \multicolumn{1}{c}{\begin{tabular}[c]{@{}c@{}}Brier\end{tabular}} & \multicolumn{1}{c}{\begin{tabular}[c]{@{}c@{}}AUROC\end{tabular}} & \multicolumn{1}{c@{}}{\begin{tabular}[c]{@{}c@{}}BA\end{tabular}} \\
\midrule
\multicolumn{5}{@{}l}{\textit{SQuAD}} \\
Qwen & native & 0.088 & 0.947 & 0.888 \\
 & empty & 0.106 & 0.923 & 0.864 \\
Gemma & native & 0.099 & 0.934 & 0.867 \\
 & empty & 0.121 & 0.904 & 0.823 \\
\midrule
\multicolumn{5}{@{}l}{\textit{RepLiQA}} \\
Qwen & native & 0.028 & 0.996 & 0.916 \\
 & empty & 0.047 & 0.984 & 0.853 \\
Gemma & native & 0.042 & 0.992 & 0.868 \\
 & empty & 0.064 & 0.978 & 0.792 \\
\bottomrule
\end{tabular}

\caption{Logistic readouts without output scores, after common probability
calibration, on the same inputs as Table~\ref{tab:primary}.
BA uses a probability threshold of 0.5.}
\label{tab:controls}
\end{table}

\subsection{\frank{RQ2: Does the Advantage Include Better Ranking?}}
\label{sec:ranking}
\frank{A Brier improvement alone does not establish better ranking.
We therefore compare AUROC for the calibrated readouts without output
scores (Table~\ref{tab:controls}).} Native training improves AUROC on
SQuAD from 0.923 to 0.947 for Qwen and from 0.904 to 0.934
for Gemma.
In all four dataset--model comparisons, adjusted post-hoc paired-cluster
intervals exclude zero in favor of native training (Appendix~\ref{app:auroc-uncertainty},
Table~\ref{tab:auroc-uncertainty}).
A strictly increasing scalar recalibration of these final probabilities
would preserve their AUROC difference. The advantage thus includes a
ranking difference that threshold adjustment alone cannot explain.
On SQuAD, these values describe the
six-component average rather than a single linear direction.

The descriptive CORP decomposition places most of the Brier gap
between readouts in discrimination (DSC). For SQuAD Qwen, DSC is 0.167
with native and 0.152 with empty training, whereas miscalibration (MCB)
differs by 0.002. \frank{DSC accounts for most of the gap in the other
three comparisons}
(Appendix~\ref{app:decomposition}).

\subsection{\frank{RQ3: How Does Native Fitting Affect Answer Retention?}}
\label{sec:answers}
Better answerability predictions need not improve the answers a system
delivers. In a separate evaluation on the official SQuAD 2.0 development
set, we vary the veto signal while keeping each model's generated
candidates fixed. The questions come from different articles than those
used for the probability comparisons.
Native readout policies improve F1 over empty readout policies by 3.99 points for Qwen
and 1.93 for Gemma (Table~\ref{tab:qa}).
Both adjusted confidence intervals exclude zero.
The original complete-output requirement remains unmet because some
generations hit the token limit, but worst-case bounds keep both
native--empty differences positive.

\begin{table}[!b]
\centering\small
\begin{tabular}{@{}lcc@{}}
\toprule
\multicolumn{1}{@{}l}{\begin{tabular}[c]{@{}l@{}}Policy\end{tabular}} & \multicolumn{1}{c}{\begin{tabular}[c]{@{}c@{}}Qwen\end{tabular}} & \multicolumn{1}{c@{}}{\begin{tabular}[c]{@{}c@{}}Gemma\end{tabular}} \\
\midrule
\multicolumn{3}{@{}l}{\textit{Answer F1}} \\
Unchanged & 81.51 & 70.65 \\
Empty readout & 78.71 & 76.92 \\
Native readout & 82.70 & 78.85 \\
Selected output & 81.67 & 78.12 \\
\midrule
Native fusion & 83.82 & 79.69 \\
Empty fusion & 83.16 & 78.23 \\
\midrule
\multicolumn{3}{@{}l}{\textit{Native readout minus comparator}} \\
Empty readout & 3.99 [2.43, 5.77] & 1.93 [0.91, 2.88] \\
Selected output & 1.03 [0.13, 1.99] & 0.72 [$-$0.06, 1.41] \\
Unchanged & 1.19 [$-$0.24, 2.62] & 8.20 [6.94, 9.57] \\
\bottomrule
\end{tabular}

\caption{Answer F1 on the official SQuAD 2.0 development set (0--100).
Policies share generated candidates within each model. Brackets give paired
article-bootstrap intervals, Bonferroni-adjusted for six comparisons at
nominal 95\% family confidence.
F1 values are lower bounds with zero credit for unfinished generations
(sensitivity bounds in \frank{Appendix~\ref{app:missing-outputs}}). Differences use unrounded scores.}
\label{tab:qa}
\end{table}

\begin{figure}[!t]
\centering
\includegraphics[width=\columnwidth]{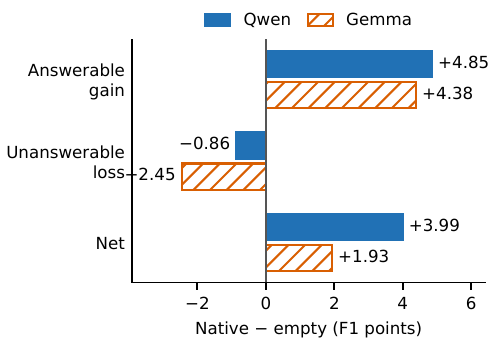}
\caption{\frank{Native-minus-empty answer F1, separated into official
credit differences on answerable and unanswerable questions.
Bars sum to the net gain over all 11,873 questions per model, with zero
credit for unfinished generations. These are arithmetic contributions
(Appendix~\ref{app:f1-contributions}).}}
\label{fig:tradeoff}
\end{figure}

The F1 gains come from retaining more correct answers, at the cost of more
answers to unanswerable questions. In both models, the additional
credit on answerable questions exceeds the loss on unanswerable questions
(Figure~\ref{fig:tradeoff}).

Both the input signal and the threshold objective affect the size of the F1 gain.
Adding the output score narrows the native--empty F1 difference to 0.67 for Qwen
and 1.46 for Gemma. With readouts alone, the F1-selected thresholds and two
alternative threshold objectives all favor native training in point
estimates, but the gains range from 0.74 to 3.99 for Qwen and 1.93 to 5.91
for Gemma. The alternatives limit retention on unanswerable calibration
questions to 5\% or 10\%, and have no paired intervals
(Appendix~\ref{app:qa}, Table~\ref{tab:all-policies}).

Against the other prespecified comparators, native readout vetoes improve
Gemma F1 over unchanged behavior by 8.20 points and Qwen F1 over the
selected output veto by 1.03 points, with both adjusted intervals above
zero (Table~\ref{tab:qa}). For Qwen, the empty readout policy scores 78.71 F1, below unchanged behavior
at 81.51, so part of the native--empty gain reflects a weak baseline.
Qwen's gain over the selected output comparator falls from 1.03 to 0.97
points when unfinished generations receive their worst-case contributions,
below the prespecified 1-point practical criterion.
A Qwen veto with isotonic output calibration is also closer to native
than the comparator selected by calibration F1 (Appendix~\ref{app:qa}).
The adjusted intervals leave the native policy's advantage over unchanged
behavior unresolved for Qwen and its advantage over output-based vetoes
unresolved for Gemma.

\subsection{\frank{RQ4: What Limits the Native Fitting Advantage?}}
\label{sec:boundaries}
We vary estimator and depth in the primary fusion comparison to test
whether the preference for native negatives extends beyond the selected
logistic readouts.
At Qwen's final SQuAD block, native logistic fusion is slightly worse
than empty fusion, by 0.0008 Brier relative to empty's loss of 0.095.
At Gemma's selected block, fusion using DiM favors empty training, by 0.0033
relative to empty's loss of 0.108.
Table~\ref{tab:estimator-depth} (Appendix~\ref{app:estimator-depth})
lists all combinations. On SQuAD, only the selected logistic contrasts
have paired intervals.
On RepLiQA, DiM favors empty training at Gemma's selected block and at
both models' final blocks, with all three exploratory 95\% intervals
excluding zero. Qwen's selected DiM fit favors native training, also with
an exploratory interval excluding zero. Final-block logistic fits favor native training
in both models, less strongly than the selected-block fits.
\frank{These reversals limit the preference for native fitting to the
tested estimator and depth.}

Simple text features predict RepLiQA labels well, raising the shortcut
concern also studied in controlled NLI evaluations
\citep{mccoy-etal-2019-right}.
Question-only AUROC is 0.982, and a classifier using question and passage
word counts reaches BA 0.953 with native training
(Appendix~\ref{app:text-controls}), above both native readouts in
Table~\ref{tab:controls}.
The word-count classifier has lower negative-class Brier (0.050) than the Qwen readout
(0.098) or Gemma readout (0.138), while readouts perform better on positives.
With roughly four positives per negative, aggregate Brier can favor
a readout that is worse on unanswerable questions.
Giving the two classes equal weight also favors the word-count classifier
over both native readouts
(Appendix~\ref{app:text-controls}).

Negative construction affects the word-count classifier too: its RepLiQA
Brier improves from 0.099 with empty training to 0.041 with native training.
The construction effect on RepLiQA is therefore not specific to
hidden-state readouts.
On the same new SQuAD articles, the previously fitted text baselines have
BA 0.506--0.543 (Appendix~\ref{app:text-controls}).
Their weaker performance makes SQuAD the more informative setting
for the hidden-state comparison.

\subsection{\frank{Additional Analysis: Threshold Transfer}}
\label{sec:threshold-results}
\frank{The QA ranking gap survives probability calibration, but a well-ordered local
signal can still fail to transfer its decision threshold. We examine that
separate issue in a retrospective constructed factual task.}

\paragraph{\frank{Task and calibration.}}
\label{sec:constructed-setup}
We adapt BBQ scenarios \citep{parrish-etal-2022-bbq} into factual questions
with three answer options, including \emph{unknown}
(Appendix~\ref{app:constructed}).
We extend a bare background (Bkg) with evidence (E) or an irrelevant
sentence (I), labeling only E as answerable. For each model, E and I
prompts are matched in total token count. E/Bkg and E/I denote the two
training contrasts.

We fit readouts by leaving out one of 8 development scenarios and
evaluate on 24 different scenarios. Notebook (I-N) and bag (I-B) variants
check wording sensitivity. Qwen block 7 and Gemma block 11, at the last
prompt token, were fixed before QA depth selection.

Training thresholds maximize BA on the fitting examples.
\frank{Common threshold calibration maximizes BA on one 12-scenario half
and tests on the other, then swaps roles, keeping readout weights fixed.
Within-scenario AUROC compares E and I variants locally, unlike pooled
QA AUROC. Metrics average across scenarios and dependent fits.
Reused templates and AI-assisted review limit generalization
(Appendix~\ref{app:constructed}).}

\paragraph{\frank{Results.}}
E/Bkg logistic readouts rank evidence above irrelevant
additions almost perfectly within each scenario, yet their training
thresholds yield BA of only 0.638 for Qwen and 0.589 for Gemma.
Common threshold calibration raises BA to 0.898 and 0.920, respectively
(Table~\ref{tab:threshold}). Strong local ranking therefore does not
ensure that a threshold transfers to new scenarios.

\begin{table}[!b]
\centering\small
\setlength{\tabcolsep}{3.5pt}
\begin{tabular}{@{}llrrr@{}}
\toprule
\multicolumn{1}{@{}l}{\begin{tabular}[c]{@{}l@{}}Model\end{tabular}} & \multicolumn{1}{l}{\begin{tabular}[c]{@{}l@{}}Training\end{tabular}} & \multicolumn{1}{c}{\begin{tabular}[c]{@{}c@{}}Within\\scenario\\AUROC\end{tabular}} & \multicolumn{1}{c}{\begin{tabular}[c]{@{}c@{}}BA\\training\\threshold\end{tabular}} & \multicolumn{1}{c@{}}{\begin{tabular}[c]{@{}c@{}}BA\\recalibrated\\threshold\end{tabular}} \\
\midrule
Q & Logistic / E/Bkg & 1.000 & 0.638 & 0.898 \\
Q & Logistic / E/I-N & 1.000 & 0.766 & 0.969 \\
Q & Logistic / E/I-B & 1.000 & 0.770 & 0.970 \\
\midrule
G & Logistic / E/Bkg & 1.000 & 0.589 & 0.920 \\
G & Logistic / E/I-N & 1.000 & 0.682 & 0.935 \\
G & Logistic / E/I-B & 1.000 & 0.738 & 0.941 \\
\bottomrule
\end{tabular}

\caption{Threshold transfer with fixed logistic readouts (Q: Qwen, G: Gemma).
Training contrasts are defined in \S\ref{sec:constructed-setup}.
Both BA columns test held-out scenarios with training thresholds or
thresholds calibrated on other scenarios. Metrics average over 24 scenarios
and dependent fits. BA includes both irrelevant-wording variants.
Appendix~\ref{app:constructed}, Table~\ref{tab:threshold-full}, also reports DiM.}
\label{tab:threshold}
\end{table}

Better thresholds account for much of the original decision gap, but
the remaining differences favor training against irrelevant additions.
After threshold calibration, E/I readouts still have higher BA point
estimates than E/Bkg readouts across both models, both estimators,
and both irrelevant wordings, giving 8 comparisons
(Appendix~\ref{app:constructed}, Table~\ref{tab:threshold-differences}).
Of the 8 conditional 95\% scenario-bootstrap intervals, 6 exclude zero,
but the 2 Gemma logistic differences, 0.015 and 0.021, remain uncertain.
These are dependent comparisons, and crossing zero does not establish equivalence.
Context-only unigram baselines exceed 0.989 BA after threshold calibration,
limiting what these decisions establish about contextual understanding.

\section{Discussion}
\frank{Evaluating how readouts rank answerable inputs against native
negatives helps determine what threshold adjustment can repair, even
when validation against empty negatives is near perfect
(Appendix~\ref{app:depth-selection}).
Our comparison identifies a limit of the tested fitting and calibration
procedures, leaving other ways of learning from native labels untested.}

\frank{Threshold adjustment can improve decisions when ranking is accurate
\citep{lavi2026directions}. Our constructed task shows this benefit across
scenarios, although near-perfect within-scenario AUROC does not establish
accurate pooled ranking.}

Choosing training negatives also requires validation with the intended
fitting procedure. \frank{The reversals across estimators and depths
(\S\ref{sec:boundaries}) prevent a uniform recommendation for native
negatives.} Our
logistic and DiM procedures differ in preprocessing as well as fitting
objective, so their contrast does not isolate the source of that
estimator dependence.

On RepLiQA, text shortcuts \citep{geirhos2020shortcut} and class imbalance
make text controls and class-conditional losses necessary to interpret
aggregate gains. \frank{The construction effect also appears in the word-count
classifier, so this dataset does not establish that the benefit requires
hidden states.}

Even a predictive advantage on native inputs is insufficient to choose
a readout for answer selection. The answer-level evaluation does not
establish consistent F1 advantages
over output-based vetoes or unchanged behavior across both models
(\S\ref{sec:answers}). Both should remain baselines for evaluating a
veto policy. \frank{Under our F1 objective, the credit from retaining
additional correct answers exceeds the loss from additional answers to
unanswerable questions.} An application
that penalizes those errors more heavily may favor another threshold.

\section{Conclusion}
\frank{At the selected depths, native-trained logistic readouts
achieve lower Brier loss and higher AUROC than empty-trained readouts
after common probability calibration. In some other estimator--depth
settings, empty fitting yields lower fusion Brier loss.
With generated candidates fixed, native readout policies
retain more correct answers and more answers to unanswerable questions,
yielding higher F1 than empty readout policies. Some comparisons with
output-based vetoes and unchanged behavior remain inconclusive.}

\label{main-text-end}
\section*{Limitations}
\noindent\textbf{\frank{Scope and identification.}}
Our evidence covers two instruction-tuned models and two English QA
datasets with different fitting protocols and class proportions.
\frank{Removing negative passages changes content, length, and text
presence together, so the comparison does not isolate their separate effects.}
The evaluation texts were unused in our earlier experiments, but
pretraining exposure or semantic overlap remains possible.

\noindent\textbf{\frank{Selection and statistical uncertainty.}}
\frank{We examine linear readouts at one token position and select a shared
depth from four candidates using native validation labels for both
training conditions. This design does not bound the best attainable
readout performance. Intervals condition on fixed fits and calibration
samples, excluding refitting and depth-selection uncertainty.}
The models also share evaluation questions, so they are not independent
replications of the sampled data.

\noindent\textbf{\frank{Shortcuts and validation.}}
\frank{The constructed factual task reuses previously inspected templates
and uses AI-assisted content review without independent human validation.}
Its lexical shortcuts, and those found in RepLiQA, limit
what high accuracy establishes about contextual understanding.
\frank{Predictive access does not identify how the model uses evidence.}

\noindent\textbf{\frank{Answer-policy evaluation.}}
Veto thresholds use 128 calibration answers per model, and the observed
F1 gap varies across operating objectives.
\frank{Unfinished generations leave the complete-output requirement unmet,
although worst-case bounds keep the native--empty F1 gains positive.}
F1 gains under the chosen thresholds establish neither a
safety improvement nor a uniform reduction in selective risk.

\section*{Ethical Considerations}
We use existing QA datasets and open-weight language models, with no
new participant data collection. The constructed factual task is adapted
from BBQ but does not measure
fairness, social bias, or harms to demographic groups.

Public source texts can contain personal names, and BBQ includes
stereotypes. To limit redistribution, the anonymous reproduction package
excludes model weights, raw passages, and generated answer text.
Appendix~\ref{app:resources} documents the artifact licenses and the
package's scope. These exclusions do not imply that the source datasets
are free of identifying or offensive content.

\paragraph{AI assistance.}
AI agents assisted with research brainstorming, code implementation, and manuscript drafting and revision. The authors reviewed the AI-assisted contributions and conducted numerical and reproducibility checks.

\bibliography{references}
\clearpage
\appendix
\section{\frank{Appendix: Data and Readout Fitting Protocol}}
\label{app:protocol}
This appendix specifies the data and fitting choices behind the controlled
comparison. It also distinguishes the diagnostics that informed the design
from evaluations on new articles and documents.

\subsection{\frank{QA Inputs and Training Negatives}}
\label{app:qa-prompts}
\frank{Both datasets request a supported answer or the prescribed
abstention response.} SQuAD requires a copied span, while RepLiQA
permits paraphrasing. The exact user-message templates are shown below,
with placeholders for the supplied text and question.

\noindent\textit{SQuAD 2.0}
\begin{quote}\small\ttfamily\raggedright
Answer using only the passage below. If the passage does not supply the answer,
output exactly UNANSWERABLE. Otherwise output only the shortest answer span
copied from the passage. Do not explain.

Passage:\\
\{context\}

Question: \{question\}\\
Answer:
\end{quote}

\noindent\textit{RepLiQA}
\begin{quote}\small\ttfamily\raggedright
Answer using only the document below. If the document does not supply the answer,
output exactly UNANSWERABLE. Otherwise give a short answer supported by the
document; paraphrasing is allowed. Do not explain.

Document:\\
\{context\}

Question: \{question\}\\
Answer:
\end{quote}

\frank{We format each template as a single user message using the model's
chat template, including the prefix for the assistant's response and no
additional system message. Empty negatives leave the passage field blank
and retain the surrounding template.}
The prompts differ across datasets but are identical across training
conditions within each dataset.
Native negatives retain the dataset's original
question and passage, whereas empty negatives retain the question and remove
the passage. Labels are never inferred from whether a model answers.
\frank{Because both conditions retain the same questions, question-only
predictors receive identical training inputs across training conditions.}
Deletion changes the input for an already labeled negative.
The label concerns passage support, not whether the model can answer
from parametric knowledge.

\subsection{\frank{Data Assignments}}
\label{app:data-selection}
\paragraph{\frank{SQuAD assignments.}}
We divide the 48-article fitting pool into three disjoint 16-article
groups. The six components use all six ordered assignments of
these groups to readout fitting, probability calibration and
internal evaluation. We average component probabilities rather than
vote on thresholded predictions. Both negative constructions and the output
comparators share the same calibration assignments.
The new-article evaluation reuses these existing fits without retraining.
We specified the six-component probability average before evaluating new articles.
In contrast, earlier held-out diagnostics averaged only the two components
for which each article belonged to the internal evaluation group.
The original 16-article calibration set is not the probability-fitting
set for these six components. It supplies the 128 generated answers used
to choose veto thresholds in the later answer-level evaluation.

\frank{To select new SQuAD articles for probability evaluation, we screen
for exact overlap with earlier inputs after normalizing passages and questions
and exclude texts shared across article titles.}
Eligible articles have at least four remaining paragraphs with both
answerable and unanswerable questions. A fixed hash ordering selects
164 articles, four paragraphs per article, and one question of each class
per paragraph, yielding 1,312 class-balanced questions.
We fix membership before model inference.
The 164-article evaluation excludes the eight
task-development articles, 48 fitting articles, 16 original calibration
articles and 32 original test articles. These checks do not guarantee
semantic independence or lack of pretraining exposure.
The official development split is used only for the later answer-level
evaluation, separately from the probability evaluation.

\paragraph{\frank{RepLiQA assignments.}}
We use the first released split, \texttt{repliqa\_0}, at dataset revision
\nolinkurl{bc880adc948fd3a70d5f8b2b3a1d1ee90d820dbd}.
We retain this version for the within-dataset comparison, rather than
benchmark the latest release. A fixed hash ordering of its 3,591 document
IDs assigns the first 24 documents to development, the next 96 to fitting,
the next 64 to calibration, and the next 256 to evaluation.
We keep all five questions per document, without sorting by labels,
length, topic, or model outcomes. The evaluation set
contains 1,020 answerable and 260 unanswerable questions.
We choose the shared depth
on fitting folds, then fit each readout on all 96 documents.
Using Brier loss across three grouped calibration folds, we select among
isotonic, scalar logistic, and per-marker-token logistic output calibration
and refit the selected map on all 64 calibration documents:
scalar logistic for Qwen and per-marker-token logistic for Gemma.
All three output alternatives remain in the saved results.

\subsection{\frank{Readout and Probability Fitting}}
\label{app:estimators}
Fits and text baselines use scikit-learn \citep{pedregosa2011sklearn}.
QA logistic fits use training-only coordinate standardization, L2
regularization with \frank{inverse regularization strength} $C=1$,
balanced class weights, \frank{and the LIBLINEAR solver. Optimization uses
at most 2,000 iterations and a random seed of 0.} \frank{For condition $t$, DiM uses the
positive and negative fitting-set activation means
$\mu_{+,t},\mu_{-,t}\in\mathbb{R}^{d_\ell}$ in raw coordinates:}
\[
w_t=\frac{\mu_{+,t}-\mu_{-,t}}{\|\mu_{+,t}-\mu_{-,t}\|_2},
\qquad b_t=0.
\]
The scalar calibration map subsequently supplies an intercept.
Probability calibration standardizes its one or two input scores using
calibration data only. Logistic calibration uses L2 regularization with
$C=1$ \frank{and an intercept. We use the LIBLINEAR solver with at most
2,000 iterations and a random seed of 0.}
For probability calibration, SQuAD uses balanced class weights, which
equal 1 on its balanced sets, whereas RepLiQA uses unweighted fitting
at its observed prevalence.
Isotonic output calibration is \frank{nondecreasing} with clipped extrapolation.

\subsection{\frank{Activation and Output Extraction}}
\label{app:extraction}
Extraction uses pinned model revisions, bfloat16 (BF16) inference and
scaled dot-product attention (SDPA). The state is the raw block output
at the last prompt token.
\frank{The output score is the negative sum of the log probabilities of the
prescribed abstention response, without an end-of-sequence token.}
For \frank{response} tokens $a_1,\ldots,a_K$, the per-marker-token comparator
uses the vector of conditional log probabilities
$\log P(a_k\mid x_i,a_{<k})$, one feature per token. RepLiQA uses these
log probabilities directly, while the SQuAD output controls negate each
feature. Both fit unconstrained logistic coefficients after standardization,
so this sign convention does not restrict the predictor family.
Tokenization checks ensure that the continuation preserves the prompt
prefix. \frank{Reference inputs verify compatibility with previously saved
states and scores before we evaluate new inputs.} We save predictions, fitted
preprocessors and weights for replay. Because refitting can change
tied TF-IDF feature selection, we report it separately from replay.

\subsection{\frank{Depth Selection and Study History}}
\label{app:depth-selection}
Both QA datasets use the same four candidate blocks per model and
three article/document-grouped fitting folds. The zero-based candidate
blocks are 7, 13, 20 and 27 for Qwen, and 11, 20, 30 and 41 for Gemma.
The selector averages native-validation logistic AUROC over folds and
both training conditions, with ties favoring the shallower block.
DiM uses the same selected depth rather than selecting its own.
For empty-trained logistic readouts, three-fold cross-validation on the
original SQuAD fitting articles yields fold-mean AUROC of
0.998--1.000 against empty negatives but 0.534--0.924 against native negatives
across eight model-by-depth settings. Near saturation against empty negatives gives
little guidance about which depth will work on native negatives.
An earlier SQuAD diagnostic also examined the state after processing
the first generated token. That position has access to a model output
and is not used in the prompt-only evaluations on new articles and documents.

\paragraph{\frank{Exploratory development.}}
Initial SQuAD readouts used Qwen block 7 and Gemma block 11, the depths
previously fixed for the constructed factual task. They performed poorly
on native inputs. Empty-trained fits reached BA 1.000 when tested against
empty negatives but 0.500 against native negatives at their matched thresholds.
Revisiting the already inspected 32-article test after fitting-only
depth selection yielded much stronger readouts. Common threshold calibration
removed or reversed some BA differences in those particular fits.
These exploratory observations motivated the subsequent new-article
design but do not replace its prespecified tests.

\section{\frank{Appendix: Probability Results and Statistical Inference}}
\label{app:probability-results}
\subsection{\frank{Primary Probability Comparisons}}
\label{app:brier-inference}
The QA probability tables report pooled AUROC and BA over all evaluation
questions. BA averages the positive- and negative-class recall after
predicting answerable when $p(x_i)>0.5$, using the averaged probability
for SQuAD. This fixed probability threshold is distinct from the F1-selected
veto thresholds used in the answer-level evaluation.

\frank{In RepLiQA,} one hundred evaluation documents contain only one class.
Their within-document AUROC/BA is undefined, not zero, but all their \frank{questions}
remain in loss estimates and cluster resampling.

\frank{For the primary Brier comparisons, Brier-loss differences
average paired cluster losses.}
SQuAD contributes eight equally weighted questions per article and
RepLiQA five per document. Each dataset's four primary differences use
two-sided 98.75\% $t$ intervals and 10,000 paired cluster bootstrap
draws. Models share the same clusters.
All intervals condition on the \frank{fitted predictors and calibration maps}.
They neither account for all historical hypothesis selection nor
provide distribution-free guarantees.

Table~\ref{tab:confirmation} reports the absolute Brier losses and
paired differences underlying Figure~\ref{fig:probability-effects}.
\begin{table*}[!t]
\centering\small
\begin{tabular}{@{}lrrrrr@{}}
\toprule
\multicolumn{1}{@{}l}{\begin{tabular}[c]{@{}l@{}}Dataset / model\end{tabular}} & \multicolumn{1}{c}{\begin{tabular}[c]{@{}c@{}}Native\\fusion\end{tabular}} & \multicolumn{1}{c}{\begin{tabular}[c]{@{}c@{}}Empty\\fusion\end{tabular}} & \multicolumn{1}{c}{\begin{tabular}[c]{@{}c@{}}Output\end{tabular}} & \multicolumn{1}{c}{\begin{tabular}[c]{@{}c@{}}Empty fusion $-$ native\\\frank{98.75\% interval}\end{tabular}} & \multicolumn{1}{c@{}}{\begin{tabular}[c]{@{}c@{}}Output $-$ native\\\frank{98.75\% interval}\end{tabular}} \\
\midrule
SQuAD / Qwen & 0.085 & 0.095 & 0.097 & 0.010 [0.005, 0.015] & 0.011 [0.006, 0.017] \\
SQuAD / Gemma & 0.095 & 0.110 & 0.119 & 0.015 [0.010, 0.019] & 0.023 [0.015, 0.031] \\
RepLiQA / Qwen & 0.029 & 0.051 & 0.066 & 0.022 [0.017, 0.027] & 0.036 [0.028, 0.044] \\
RepLiQA / Gemma & 0.043 & 0.064 & 0.073 & 0.021 [0.017, 0.026] & 0.030 [0.024, 0.036] \\
\bottomrule
\end{tabular}

\caption{Primary Brier comparisons on new SQuAD articles and
RepLiQA documents. Lower Brier is better. Difference columns subtract
native-fusion Brier from empty-fusion or output Brier, so positive values
favor native fusion. Intervals are paired 98.75\% article/document $t$ intervals,
corrected for four comparisons per dataset. The datasets use different
fitting protocols and prevalences, so effects are not pooled.
Output is the prespecified comparator. Differences are computed before rounding.}
\label{tab:confirmation}
\end{table*}

\subsection{\frank{Post-hoc Ranking Uncertainty}}
\label{app:auroc-uncertainty}
After examining the AUROC point estimates, we computed paired cluster
bootstrap intervals for the four selected-depth logistic readout-alone
differences in Table~\ref{tab:auroc-uncertainty}. This supplements the
prespecified Brier tests rather than adding a new confirmation study.
We kept all predictors, calibration maps, and evaluation inputs fixed.
For SQuAD, predictions remain the mean of six calibrated components.

We drew 20,000 samples of 164 articles for SQuAD and 256 documents for
RepLiQA, sampling clusters with replacement and retaining every question
with its cluster's multiplicity. Both training conditions and both models
share the draws within each dataset. Each draw recomputes pooled AUROC,
with half credit for tied scores, rather than averaging within-cluster
AUROCs. We retained RepLiQA's 100 single-class documents. Every resampled
pool contained both classes.

The table reports ordinary 95\% percentile intervals and Bonferroni-adjusted
98.75\% intervals for these four supplementary contrasts across both
datasets. This family is separate from the four primary Brier comparisons
within each dataset. We sorted cluster identifiers and used seeds
20260928 and 20260929 for SQuAD and RepLiQA, respectively.
The intervals quantify evaluation-sampling uncertainty conditional on
the fitted predictors. They do not account for refitting, depth selection,
or the earlier inspection of these results.

\begin{table}[!tb]
\centering\footnotesize
\setlength{\tabcolsep}{2pt}
\begin{tabular}{@{}lrrr@{}}
\toprule
\multicolumn{1}{@{}l}{\begin{tabular}[c]{@{}l@{}}Model\end{tabular}} & \multicolumn{1}{c}{\begin{tabular}[c]{@{}c@{}}$\Delta$AUROC\end{tabular}} & \multicolumn{1}{c}{\begin{tabular}[c]{@{}c@{}}\frank{95\% interval}\end{tabular}} & \multicolumn{1}{c@{}}{\begin{tabular}[c]{@{}c@{}}Adjusted\\\frank{98.75\% interval}\end{tabular}} \\
\midrule
\multicolumn{4}{@{}l}{\textit{SQuAD}} \\
Qwen & 0.0237 & [0.0141, 0.0337] & [0.0115, 0.0368] \\
Gemma & 0.0293 & [0.0180, 0.0412] & [0.0153, 0.0450] \\
\midrule
\multicolumn{4}{@{}l}{\textit{RepLiQA}} \\
Qwen & 0.0119 & [0.0061, 0.0188] & [0.0049, 0.0212] \\
Gemma & 0.0149 & [0.0090, 0.0217] & [0.0077, 0.0240] \\
\bottomrule
\end{tabular}

\caption{Post-hoc paired-cluster intervals for the readout-alone AUROCs
in Table~\ref{tab:controls}. $\Delta$ is native minus empty. Intervals condition
on fixed predictors and resample 164 SQuAD articles or 256 RepLiQA documents.
Adjustment covers these four supplementary contrasts, all with positive
lower bounds.}
\label{tab:auroc-uncertainty}
\end{table}

\subsection{\frank{Descriptive Loss Decomposition}}
\label{app:decomposition}
We use the consistent, optimally binned, reproducible, and
pool-adjacent-violators-based (CORP) decomposition to describe the Brier-loss gap
after common probability calibration. \frank{We keep the evaluated probabilities fixed and fit a diagnostic
isotonic map to evaluation labels using the pool-adjacent-violators
(PAV) algorithm.}
Let $\mathrm{BS}_{\mathrm{iso}}$ denote the loss after isotonic
recalibration of the evaluated probabilities, and let
$\mathrm{BS}_{\mathrm{prev}}$ be the empirical prevalence predictor's loss.
\frank{The empirical miscalibration (MCB), discrimination (DSC), and
uncertainty (UNC) terms are
$\mathrm{MCB}=\mathrm{BS}-\mathrm{BS}_{\mathrm{iso}}$,
$\mathrm{DSC}=\mathrm{BS}_{\mathrm{prev}}-\mathrm{BS}_{\mathrm{iso}}$,
and $\mathrm{UNC}=\mathrm{BS}_{\mathrm{prev}}$, giving}
\begin{equation}
\mathrm{BS}=\mathrm{MCB}-\mathrm{DSC}+\mathrm{UNC}.
\label{eq:corp}
\end{equation}
The evaluation-label isotonic fit is descriptive, not a deployed correction.

\begin{table}[!htbp]
\centering\small
\setlength{\tabcolsep}{4pt}
\begin{tabular}{@{}lrrrr@{}}
\toprule
\multicolumn{1}{@{}l}{\begin{tabular}[c]{@{}l@{}}Model / training\end{tabular}} & \multicolumn{1}{c}{\begin{tabular}[c]{@{}c@{}}BS\end{tabular}} & \multicolumn{1}{c}{\begin{tabular}[c]{@{}c@{}}MCB\end{tabular}} & \multicolumn{1}{c}{\begin{tabular}[c]{@{}c@{}}DSC\end{tabular}} & \multicolumn{1}{c@{}}{\begin{tabular}[c]{@{}c@{}}UNC\end{tabular}} \\
\midrule
\multicolumn{5}{@{}l}{\textit{SQuAD}} \\
Qwen / native & 0.088 & 0.006 & 0.167 & 0.250 \\
Qwen / empty & 0.106 & 0.008 & 0.152 & 0.250 \\
Gemma / native & 0.099 & 0.005 & 0.156 & 0.250 \\
Gemma / empty & 0.121 & 0.007 & 0.136 & 0.250 \\
\midrule
\multicolumn{5}{@{}l}{\textit{RepLiQA}} \\
Qwen / native & 0.028 & 0.013 & 0.147 & 0.162 \\
Qwen / empty & 0.047 & 0.018 & 0.132 & 0.162 \\
Gemma / native & 0.042 & 0.020 & 0.140 & 0.162 \\
Gemma / empty & 0.064 & 0.024 & 0.123 & 0.162 \\
\bottomrule
\end{tabular}

\caption{Empirical CORP decomposition of fixed readout-alone
probabilities on new articles and documents. Isotonic PAV is fitted on evaluation
labels only for this descriptive analysis. BS = MCB $-$ DSC + UNC,
up to displayed rounding. No PAV-adjusted prediction is deployed.}
\label{tab:decomposition}
\end{table}

The decomposition uses no arbitrary probability bins. Its empirical
PAV fit can be optimistic on finite evaluation samples, so the analysis
does not replace the out-of-sample Brier endpoint.
The uncertainty term differs across datasets because their class
proportions differ, so we compare decomposition terms within each dataset
rather than pool them across protocols. Higher DSC is not proof
of semantic evidence access or of better decisions at every threshold.

\section{\frank{Appendix: Answer Policies and Unfinished Generations}}
\label{app:qa}
The answer-level comparison asks how alternative vetoes change the same
generated candidates. We report all tested policies and explicitly retain
the uncertainty from unfinished generations.

\subsection{\frank{Candidates, Thresholds, and Prespecified Comparisons}}
The evaluation covers all 11,873 official development questions across
35 articles: 5,928 answerable and 5,945 unanswerable.
Each policy sees the same one-shot greedy candidate. A veto policy thresholds
its calibrated answerability probability to retain the candidate or return
an empty answer. It cannot force regeneration or recover an answer after the model abstains.
We select thresholds and the output comparator only on the
128-answer calibration set, choosing scalar logistic output calibration for Qwen and
isotonic output calibration for Gemma. These downstream choices differ
from the comparator definitions in the earlier Brier experiments.
\frank{The primary veto thresholds maximize answer F1 on these calibration
answers, as does selection of the output comparator.}

The six comparisons specified before development inference contain
native readout minus empty readout, selected output and unchanged
behavior in each model. We specified a gain of at least
1 F1 point and a positive multiplicity-adjusted lower bound
as practical and statistical criteria, conditional
on complete measurement. These criteria are not inferred from
the observed improvements. Fusion and \frank{policies with budgets of 0.05 or 0.10 for retaining
answers to unanswerable calibration questions} remain secondary.

\subsection{\frank{Scoring and Sensitivity to Unfinished Generations}}
\label{app:missing-outputs}
\frank{After stripping surrounding whitespace, we map responses exactly
matching the prescribed abstention response to an empty answer before
official scoring.}
Official SQuAD normalization removes case, punctuation, and articles,
and F1 maximizes over accepted reference answers. Empty normalized
reference alternatives are handled as in the official evaluator.
We verify the scorer on observable \frank{responses}, rather than treat an
unfinished generation as a known answer.
We do not selectively retry cases, increase the token limit,
or exclude cases from evaluation.

Under lower-bound scoring, \frank{a question with an unfinished generation} has zero
credit for every policy. \frank{If $m$ of $N$ generations are unfinished, each absolute
mean can lie at most $100m/N$ F1 points above its displayed value on the
0--100 scale.}
For a paired difference, allowing each unfinished \frank{generation's} contribution to
\frank{range from $-1$ to $1$} gives a deliberately loose sensitivity range.
Article-bootstrap sensitivity envelopes apply the same bounds
inside each resample. They ignore shared-candidate constraints and
do not assert certified statistical coverage.

All unfinished generations hit the 64-token limit:
seven Qwen \frank{generations} across six articles and
23 Gemma \frank{generations} across 14 articles. \frank{Of the corresponding
questions, five for Qwen and 16 for Gemma are answerable.} \frank{With these counts, each absolute mean can lie at most
0.059 points for Qwen and 0.194 for Gemma above its reported lower bound.}
The overall prespecified
complete-output requirement remains unmet even where numerical
comparison intervals exclude zero.
\frank{For native minus empty, the bootstrap sensitivity envelopes span
2.35 to 5.83 F1 points for Qwen and 0.70 to 3.05 for Gemma.
Under zero-credit scoring, the corresponding intervals span
2.43 to 5.77 and 0.91 to 2.88 points, respectively.}

\subsection{\frank{Full Policy Results}}
\frank{Table~\ref{tab:qa-intervals} reports the primary paired differences
and their uncertainty, and Table~\ref{tab:all-policies} gives all tested
operating points under the same lower-bound scoring rule.}

\begin{table*}[t]
\centering\small
\begin{tabular}{@{}lrrr@{}}
\toprule
\multicolumn{1}{@{}l}{\begin{tabular}[c]{@{}l@{}}Native minus\end{tabular}} & \multicolumn{1}{c}{\begin{tabular}[c]{@{}c@{}}F1 gain\end{tabular}} & \multicolumn{1}{c}{\begin{tabular}[c]{@{}c@{}}\frank{Adjusted interval}\end{tabular}} & \multicolumn{1}{c@{}}{\begin{tabular}[c]{@{}c@{}}Bounds for\\unfinished generations\end{tabular}} \\
\midrule
Qwen / empty & 3.99 & [2.43, 5.77] & [3.93, 4.05] \\
Qwen / output & 1.03 & [0.13, 1.99] & [0.97, 1.09] \\
Qwen / unchanged & 1.19 & [$-$0.24, 2.62] & [1.13, 1.25] \\
Gemma / empty & 1.93 & [0.91, 2.88] & [1.74, 2.13] \\
Gemma / output & 0.72 & [$-$0.06, 1.41] & [0.53, 0.92] \\
Gemma / unchanged & 8.20 & [6.94, 9.57] & [8.01, 8.39] \\
\bottomrule
\end{tabular}

\caption{Primary answer-F1 differences in percentage points.
Intervals use 20,000 paired article-bootstrap draws and a Bonferroni
adjustment for six comparisons (99.1667\% per interval for a nominal
95\% family level). The last column allows each paired difference with
an unfinished generation
\frank{to range independently from $-1$ to $1$}, a conservative point sensitivity
bound. It is not a confidence interval or a replacement for the
unmet complete-output requirement. Differences are computed before rounding.}
\label{tab:qa-intervals}
\end{table*}

\begin{table}[!t]
\centering\small
\begin{tabular}{@{}lrr@{}}
\toprule
\multicolumn{1}{@{}l}{\begin{tabular}[c]{@{}l@{}}Policy\end{tabular}} & \multicolumn{1}{c}{\begin{tabular}[c]{@{}c@{}}Qwen F1\end{tabular}} & \multicolumn{1}{c@{}}{\begin{tabular}[c]{@{}c@{}}Gemma F1\end{tabular}} \\
\midrule
Unchanged & 81.51 & 70.65 \\
Always abstain & 50.05 & 50.01 \\
Native readout & 82.70 & 78.85 \\
Native fusion & 83.82 & 79.69 \\
Output (isotonic) & 82.41 & 78.12\textsuperscript{sel} \\
Output (logistic) & 81.67\textsuperscript{sel} & 78.12 \\
Empty readout & 78.71 & 76.92 \\
Empty fusion & 83.16 & 78.23 \\
Native readout / budget 0.05 & 79.45 & 65.71 \\
Native readout / budget 0.10 & 83.13 & 72.93 \\
Native fusion / budget 0.05 & 83.82 & 60.19 \\
Native fusion / budget 0.10 & 84.18 & 71.80 \\
Output (isotonic) / budget 0.05 & 79.18 & 59.39 \\
Output (isotonic) / budget 0.10 & 82.41 & 62.79 \\
Output (logistic) / budget 0.05 & 80.89 & 61.06 \\
Output (logistic) / budget 0.10 & 82.88 & 64.93 \\
Empty readout / budget 0.05 & 78.71 & 61.31 \\
Empty readout / budget 0.10 & 80.85 & 67.01 \\
Empty fusion / budget 0.05 & 81.26 & 61.21 \\
Empty fusion / budget 0.10 & 83.16 & 70.55 \\
\bottomrule
\end{tabular}

\caption{All evaluated policy F1 scores (0--100), under the same
lower-bound convention. Superscript \textsuperscript{sel} marks the
output comparator selected by calibration answer F1 for each model
and reported as Selected output in Table~\ref{tab:qa}.
Budget denotes the allowed fraction of unanswerable calibration questions
whose answers are retained, not a certified test-risk bound. These
thresholds are secondary operating points, not alternatives selected
after viewing development-set results.}
\label{tab:all-policies}
\end{table}

Under zero-credit scoring, native has higher F1 in 32 of 35 articles
for Qwen and 26 of 35 for Gemma. These article counts describe the
spread of the effect, whereas the overall F1 endpoint remains question-weighted.

\frank{At a secondary operating point, Gemma's native-fusion policy uses a
budget of 0.05 for retaining answers to unanswerable calibration questions. It suppresses 3,078
exactly correct positive answers and scores 60.19 F1, below unchanged
behavior's 70.65.} A low retention budget need not improve
the chosen answer utility.

\subsection{\frank{F1 Contributions}}
\label{app:f1-contributions}
\noindent\frank{For question $i$, let $\Delta_i$ be the native policy's official
F1 credit minus the empty policy's credit on the 0--1 scale.
With $N=11{,}873$ questions, the native-minus-empty difference in F1 points is}
\[
\frank{\Delta\mathrm{F1}=\frac{100}{N}\left(
\sum_{i:y_i=1}\Delta_i+\sum_{i:y_i=0}\Delta_i\right).}
\]
\frank{Multiplying each sum by $100/N$ gives the answerable- and
unanswerable-question contributions in Figure~\ref{fig:tradeoff}. Questions with unfinished generations have
$\Delta_i=0$ under the reporting convention and remain in $N$.
For an observable unanswerable question, vetoing a candidate that
previously received zero credit recovers one unit of F1 credit.
Rejection counts describe that case, whereas the exact decomposition
uses official credits, including normalization and partial matches.
It describes changes in the delivered answers, not causal effects of
learned representations.}
For Qwen, the native policy rejects 762 answers to unanswerable questions,
compared with 864 under the empty policy, but rejects only 558 exactly
correct answers, compared with 1,060.
For Gemma, the native policy rejects 1,159 answers to unanswerable
questions compared with 1,450 under the empty policy, while rejecting
157 exactly correct answers compared with 587. These counts complement the partial-credit contributions
shown in Figure~\ref{fig:tradeoff}.

\section{\frank{Appendix: Additional QA Controls}}
\label{app:qa-controls}
\frank{These analyses test the dependence of the QA results on output-feature
and calibration-label access, estimator and depth, and simple text cues.
They supplement the probability comparisons in Appendix~\ref{app:probability-results}.}

\subsection{\frank{SQuAD Output Features and Calibration Access}}
\label{app:output-controls}
\frank{Table~\ref{tab:output-controls} includes the original isotonic
comparator and post-hoc alternatives that test output features and
calibration-label access. We specified the additional maps after inspecting
the new-article results and fitted them using only the original fitting-pool labels.}
The 32-article versions join a component's readout-fitting and
probability-calibration groups. All six components still share the
original 48-article pool.

\begin{table}[!htbp]
\centering\small
\begin{tabular}{@{}lrr@{}}
\toprule
\multicolumn{1}{@{}l}{\begin{tabular}[c]{@{}l@{}}Output calibration\\/ articles\end{tabular}} & \multicolumn{1}{c}{\begin{tabular}[c]{@{}c@{}}Qwen BS\end{tabular}} & \multicolumn{1}{c@{}}{\begin{tabular}[c]{@{}c@{}}Gemma BS\end{tabular}} \\
\midrule
Scalar logistic / 16 & 0.107 & 0.119 \\
Isotonic / 16 & 0.097 & 0.119 \\
Per-marker-token logistic / 16 & 0.109 & 0.109 \\
Scalar logistic / 32 & 0.106 & 0.119 \\
Isotonic / 32 & 0.097 & 0.119 \\
Per-marker-token logistic / 32 & 0.109 & 0.110 \\
\bottomrule
\end{tabular}

\caption{\frank{Original comparator and post-hoc SQuAD output controls}
evaluated on the same 164
evaluation articles. Article counts refer to each component's
calibration access, not six disjoint calibration sets.
Native fusion Brier is 0.085 for Qwen and 0.095 for Gemma.
\frank{Per-marker-token logistic uses one feature per token of the
prescribed abstention response.}}
\label{tab:output-controls}
\end{table}

Richer token features improve Gemma's
output baseline, but every tested output variant remains worse
than native fusion in Brier, with positive descriptive paired
95\% intervals. This does not cover all vocabulary logits,
alternative prompts or generated-answer features.

\subsection{\frank{Estimator and Depth Controls}}
\label{app:estimator-depth}
Table~\ref{tab:estimator-depth} completes the estimator and depth comparisons
from \S\ref{sec:boundaries}. All predictors are fusions calibrated on
the same native examples within each dataset. The table reuses saved
losses and intervals, retaining point estimates where no paired interval
was previously computed.
\begin{table*}[!t]
\centering\small
\begin{tabular}{@{}lrrrr@{}}
\toprule
\multicolumn{1}{@{}l}{\begin{tabular}[c]{@{}l@{}}Block / estimator\end{tabular}} & \multicolumn{1}{c}{\begin{tabular}[c]{@{}c@{}}Native BS\end{tabular}} & \multicolumn{1}{c}{\begin{tabular}[c]{@{}c@{}}Empty BS\end{tabular}} & \multicolumn{1}{c}{\begin{tabular}[c]{@{}c@{}}Empty $-$ native\end{tabular}} & \multicolumn{1}{c@{}}{\begin{tabular}[c]{@{}c@{}}Reported interval\end{tabular}} \\
\midrule
\multicolumn{5}{@{}l}{\textit{SQuAD / Qwen}} \\
20 (selected) / Logistic & 0.08535 & 0.09526 & +0.00991 & [0.00518, 0.01464]\textsuperscript{*} \\
20 (selected) / DiM & 0.09234 & 0.09385 & +0.00150 & --- \\
27 (final) / Logistic & 0.09561 & 0.09484 & $-$0.00077 & --- \\
27 (final) / DiM & 0.09534 & 0.09508 & $-$0.00027 & --- \\
\midrule
\multicolumn{5}{@{}l}{\textit{SQuAD / Gemma}} \\
30 (selected) / Logistic & 0.09536 & 0.10988 & +0.01453 & [0.00958, 0.01948]\textsuperscript{*} \\
30 (selected) / DiM & 0.11119 & 0.10790 & $-$0.00329 & --- \\
41 (final) / Logistic & 0.10514 & 0.10876 & +0.00362 & --- \\
41 (final) / DiM & 0.11545 & 0.11650 & +0.00105 & --- \\
\midrule
\multicolumn{5}{@{}l}{\textit{RepLiQA / Qwen}} \\
20 (selected) / Logistic & 0.02939 & 0.05118 & +0.02179 & [0.01651, 0.02707]\textsuperscript{*} \\
20 (selected) / DiM & 0.04123 & 0.04841 & +0.00718 & [0.00538, 0.00898] \\
27 (final) / Logistic & 0.04845 & 0.06064 & +0.01219 & [0.00971, 0.01467] \\
27 (final) / DiM & 0.06308 & 0.05997 & $-$0.00312 & [$-$0.00378, $-$0.00246] \\
\midrule
\multicolumn{5}{@{}l}{\textit{RepLiQA / Gemma}} \\
30 (selected) / Logistic & 0.04269 & 0.06416 & +0.02147 & [0.01708, 0.02586]\textsuperscript{*} \\
30 (selected) / DiM & 0.05102 & 0.04729 & $-$0.00373 & [$-$0.00542, $-$0.00204] \\
41 (final) / Logistic & 0.05403 & 0.06689 & +0.01286 & [0.01042, 0.01530] \\
41 (final) / DiM & 0.06523 & 0.06445 & $-$0.00078 & [$-$0.00113, $-$0.00043] \\
\bottomrule
\end{tabular}

\caption{All evaluated estimator--depth combinations on the new articles
and documents. BS is fusion Brier loss. Positive empty-minus-native
differences favor native training. Intervals are paired article/document
$t$ intervals, conditional on the fits. Asterisks mark the primary
98.75\% intervals corrected for four comparisons per dataset.
Other intervals are exploratory 95\% intervals without multiplicity
correction. A dash means no paired interval was computed, not no effect.
Five decimals preserve the signs of small differences.}
\label{tab:estimator-depth}
\end{table*}

On RepLiQA, the selected-minus-final difference in native logistic
advantage is 0.00960 for Qwen \frank{and
0.00861 for Gemma. The corresponding 95\% intervals span
0.00689 to 0.01231 and 0.00666 to 0.01056, respectively.} These are direct exploratory
comparisons of the construction effects, not a test of monotonicity
across depth. Neither dataset's controls account for uncertainty from
refitting or selection.

\subsection{\frank{Component and Original-Fit Checks}}
\label{app:component-controls}
At the selected depth, native logistic fusion has lower Brier loss
than empty logistic fusion in all six individual components,
as well as in their average.
The component gains span 0.00535--0.00881 in Qwen and 0.00344--0.01563
in Gemma. They are dependent fits, not six independent replications.

A separate replay of the original full-48-article readouts on the
new articles preserves native ranking advantages, yet Gemma's
original-threshold BA favors empty training (0.799 versus 0.829).
For Gemma, better ranking therefore coexists with lower BA at the
original thresholds.

\subsection{\frank{Text Controls}}
\label{app:text-controls}
TF-IDF uses one- and two-grams with at most 5,000 features. The simpler
word-count classifier uses two scalar features, the question and passage word counts.

Table~\ref{tab:text-controls} reports the text baselines on the same new
evaluation articles and documents as the readouts. SQuAD's original
48-article text fits use thresholds calibrated on 16 separate articles,
so their training budgets differ from the readout ensemble's component budgets.
RepLiQA text models use 96 fitting and 64 calibration documents.
Results are shared across the language models because these predictors
use no hidden states.
\begin{table*}[t]
\centering\small
\begin{tabular}{@{}lrrr@{}}
\toprule
\multicolumn{1}{@{}l}{\begin{tabular}[c]{@{}l@{}}Dataset / text predictor\end{tabular}} & \multicolumn{1}{c}{\begin{tabular}[c]{@{}c@{}}Brier\end{tabular}} & \multicolumn{1}{c}{\begin{tabular}[c]{@{}c@{}}AUROC\end{tabular}} & \multicolumn{1}{c@{}}{\begin{tabular}[c]{@{}c@{}}BA\end{tabular}} \\
\midrule
SQuAD / question only & --- & 0.558 & 0.543 \\
SQuAD / question + passage & --- & 0.510 & 0.506 \\
SQuAD / word counts, native & --- & 0.533 & 0.527 \\
RepLiQA / question only & 0.048 & 0.982 & 0.905 \\
RepLiQA / word counts, native & 0.041 & 0.986 & 0.953 \\
RepLiQA / word counts, empty & 0.099 & 0.898 & 0.753 \\
\bottomrule
\end{tabular}

\caption{Text-only controls. SQuAD entries replay native-trained
predictors under common threshold calibration, for which Brier is not
reported (---).
RepLiQA entries use probability calibration. The protocols differ
across datasets.}
\label{tab:text-controls}
\end{table*}

\paragraph{\frank{Class-conditional losses.}}
Table~\ref{tab:class-controls} gives the descriptive class-conditional
RepLiQA comparison. Class-balanced Brier loss averages positive- and negative-class
losses equally, whereas the primary endpoint retains natural prevalence.
This secondary reweighting does not replace the prespecified endpoint.
\begin{table}[!htbp]
\centering\small
\begin{tabular}{@{}lrrr@{}}
\toprule
\multicolumn{1}{@{}l}{\begin{tabular}[c]{@{}l@{}}Predictor\end{tabular}} & \multicolumn{1}{c}{\begin{tabular}[c]{@{}c@{}}Positive\\BS\end{tabular}} & \multicolumn{1}{c}{\begin{tabular}[c]{@{}c@{}}Negative\\BS\end{tabular}} & \multicolumn{1}{c@{}}{\begin{tabular}[c]{@{}c@{}}Class-balanced\\BS\end{tabular}} \\
\midrule
Word counts & 0.038 & 0.050 & 0.044 \\
Qwen / readout & 0.011 & 0.098 & 0.054 \\
Qwen / fusion & 0.011 & 0.102 & 0.056 \\
Gemma / readout & 0.017 & 0.138 & 0.078 \\
Gemma / fusion & 0.018 & 0.139 & 0.079 \\
\bottomrule
\end{tabular}

\caption{RepLiQA class-conditional Brier losses, all with native training.
The word-count classifier has lower negative-class and class-balanced loss
than both models' readouts. Under the original class prevalence,
Qwen's readout has lower overall loss than the word-count classifier, whereas Gemma's does not.}
\label{tab:class-controls}
\end{table}

\section{\frank{Appendix: Constructed-Task Diagnostics}}
\label{app:constructed}
\subsection{\frank{Construction and Provenance}}
\begin{figure*}[t]
\input{figures/constructed_example}
\caption{Actual inputs from development scenario Age-466, shown for
Qwen with the short diagram background and the first option order.
\texttt{[slot]} marks the insertion position and is not literal prompt text.
The correct option is C for E and B for Bkg, I-N, and I-B.
Including Qwen's chat template, the saved inputs contain 127 tokens for
Bkg and 143 for each of E, I-N, and I-B. These are input labels and lengths,
not model predictions.}
\label{fig:constructed-example}
\end{figure*}

The constructed factual task has eight development scenarios and
24 evaluation scenarios. Each scenario supplies
evidence and two irrelevant-addition variants, with answer rotations
and background conditions. Evaluation totals are 1,008 inputs per model:
288 evidence inputs, 288 per irrelevant variant and 144 bare backgrounds.
The combined evidence/irrelevant evaluation comprises 864 inputs.

The variants use notebook/bag sentences and model-specific token
matching. The evidence label refers to the designed task, not an
independent human annotation. Content review was AI-assisted.
The 24 scenario identifiers are disjoint from training, but templates
and components were exposed during development. Fifteen questions
change the gold answer relative to their BBQ source
\citep{parrish-etal-2022-bbq}. \frank{These results do not establish
performance on the BBQ benchmark or generalization across templates.}

Figure~\ref{fig:constructed-example} illustrates one development scenario.

\subsection{\frank{Fitting and Threshold Calibration}}
\frank{Each fit uses seven development scenarios, with 84 positive and
84 negative examples. Because these fits overlap, we do not treat them
as independent samples.}

Logistic regression
standardizes coordinates using fitting-set means and standard deviations,
then uses L2 regularization with $C=1$, the \frank{L-BFGS} solver,
a 2,000-iteration limit, and \frank{random} seed 20260917. DiM uses the normalized raw
class-mean difference and centers scores at the overall fitting mean.

For both estimators, the training threshold maximizes BA on the fitting
examples. Predictions are positive only when the score strictly exceeds
the threshold. Candidate thresholds are the distinct fitting scores and
the next representable value below their minimum, covering all-positive
and all-negative decisions. Ties in BA favor the larger threshold.

Common threshold calibration partitions scenarios into two deterministic
12-scenario halves, selects the BA-maximizing threshold on one half,
and evaluates the other before swapping their roles. We never reverse
score signs using test labels. The main text averages over the
dependent fits. Bootstrap diagnostics resample scenarios, with fits
and calibration halves held fixed.

\subsection{\frank{Results}}
Table~\ref{tab:threshold-full} gives both estimators' ranking and
decision results, extending the logistic summary in Table~\ref{tab:threshold}.
Table~\ref{tab:threshold-differences} reports the residual BA differences
after common threshold calibration. Both training conditions are evaluated
on the same scenarios, including both irrelevant-wording variants.

\begin{table}[!htbp]
\centering\small
\setlength{\tabcolsep}{3.5pt}
\begin{tabular}{@{}llrrr@{}}
\toprule
\multicolumn{1}{@{}l}{\begin{tabular}[c]{@{}l@{}}Model\end{tabular}} & \multicolumn{1}{l}{\begin{tabular}[c]{@{}l@{}}Training\end{tabular}} & \multicolumn{1}{c}{\begin{tabular}[c]{@{}c@{}}Within\\scenario\\AUROC\end{tabular}} & \multicolumn{1}{c}{\begin{tabular}[c]{@{}c@{}}BA\\training\\threshold\end{tabular}} & \multicolumn{1}{c@{}}{\begin{tabular}[c]{@{}c@{}}BA\\recalibrated\\threshold\end{tabular}} \\
\midrule
Q & DiM / E/Bkg & 0.986 & 0.819 & 0.873 \\
Q & DiM / E/I-N & 0.999 & 0.940 & 0.950 \\
Q & DiM / E/I-B & 1.000 & 0.959 & 0.965 \\
\midrule
Q & Logistic / E/Bkg & 1.000 & 0.638 & 0.898 \\
Q & Logistic / E/I-N & 1.000 & 0.766 & 0.969 \\
Q & Logistic / E/I-B & 1.000 & 0.770 & 0.970 \\
\midrule
G & DiM / E/Bkg & 0.970 & 0.570 & 0.810 \\
G & DiM / E/I-N & 0.993 & 0.874 & 0.882 \\
G & DiM / E/I-B & 0.998 & 0.919 & 0.910 \\
\midrule
G & Logistic / E/Bkg & 1.000 & 0.589 & 0.920 \\
G & Logistic / E/I-N & 1.000 & 0.682 & 0.935 \\
G & Logistic / E/I-B & 1.000 & 0.738 & 0.941 \\
\bottomrule
\end{tabular}

\caption{Complete threshold-transfer results for the constructed factual
task (Q: Qwen, G: Gemma). E/Bkg uses bare-background negatives, and E/I-N
and E/I-B use notebook and bag additions. Both BA columns evaluate
held-out scenarios, before or after threshold calibration on other scenarios.
Metrics average over 24 scenarios and dependent fits, with BA including
both irrelevant-wording variants.}
\label{tab:threshold-full}
\end{table}

\begin{table}[!htbp]
\centering\small
\begin{tabular}{@{}lrr@{}}
\toprule
\multicolumn{1}{@{}l}{\begin{tabular}[c]{@{}l@{}}Model / E/I fit\end{tabular}} & \multicolumn{1}{c}{\begin{tabular}[c]{@{}c@{}}BA gain\end{tabular}} & \multicolumn{1}{c@{}}{\begin{tabular}[c]{@{}c@{}}95\% interval\end{tabular}} \\
\midrule
Q / DiM / I-N & 0.077 & [0.031, 0.132] \\
Q / DiM / I-B & 0.092 & [0.047, 0.145] \\
Q / Logistic / I-N & 0.071 & [0.019, 0.128] \\
Q / Logistic / I-B & 0.072 & [0.021, 0.132] \\
\midrule
G / DiM / I-N & 0.072 & [0.033, 0.114] \\
G / DiM / I-B & 0.100 & [0.056, 0.150] \\
G / Logistic / I-N & 0.015 & [$-$0.026, 0.060] \\
G / Logistic / I-B & 0.021 & [$-$0.017, 0.068] \\
\bottomrule
\end{tabular}

\caption{E/I minus E/Bkg BA after common threshold calibration.
Q and G denote Qwen and Gemma. I-N and I-B denote notebook and bag
training variants. The descriptive paired 95\% scenario-bootstrap
intervals condition on fitted readouts and calibration halves,
without multiplicity correction.}
\label{tab:threshold-differences}
\end{table}

\FloatBarrier
\section{\frank{Appendix: Resources and Reproducibility}}
\label{app:resources}

\paragraph{Data and model terms.}
SQuAD~2.0 is distributed under CC BY-SA~4.0, and RepLiQA and BBQ under
CC BY~4.0.\footnote{\url{https://rajpurkar.github.io/SQuAD-explorer/},
\url{https://huggingface.co/datasets/ServiceNow/repliqa},
\url{https://github.com/nyu-mll/BBQ/blob/main/LICENSE}.}
Our factual task modifies questions and adds diagnostic text to selected
BBQ scenarios without evaluating bias.
Qwen2.5-7B-Instruct uses Apache~2.0, whereas Gemma-2-9B-it is governed
by the Gemma Terms of Use.\footnote{\url{https://huggingface.co/Qwen/Qwen2.5-7B-Instruct},
\url{https://ai.google.dev/gemma/terms}.}
The research artifacts retain attribution and license notices and
exclude model weights.

\paragraph{\frank{Model inference environment.}}
The saved extraction records for the new SQuAD articles, RepLiQA
documents and SQuAD development-set generations report NVIDIA RTX A6000
GPUs, PyTorch 2.9.1 with CUDA 12.8, Transformers 4.57.6, \frank{bfloat16
precision and scaled dot-product attention}. Qwen and Gemma have approximately 7B and 9B parameters,
respectively, and their weights are not updated.

\paragraph{\frank{CPU replay environment.}}
We tested CPU metric replay with Python 3.11.15, NumPy 2.4.6,
SciPy 1.17.1 and scikit-learn 1.9.0. These versions describe the tested
replay environment, not every historical experiment.

\paragraph{\frank{Workloads and runtime coverage.}}
The two probability evaluations process 1,312 and 1,280 inputs per
model, and the answer-level evaluation generates from 11,873 inputs
per model with a 64-token limit. These counts exclude development,
selection and earlier diagnostic runs. Complete runtime records are
unavailable for those earlier runs, so we cannot report total project GPU-hours.

\paragraph{\frank{Artifact contents and replay scope.}}
The anonymous code and data archives support CPU recomputation of the
primary Brier comparisons and paired intervals, post-hoc paired-cluster
AUROC intervals, descriptive CORP terms,
and answer-level F1 comparisons from saved probabilities and per-question
credits. They also regenerate the reported tables and result figures
from the accompanying result records.

CPU replay uses saved records, without new model inference, independent
rescoring of generated text, or refitting every historical diagnostic.
The package excludes model weights, activation caches, and raw text,
but includes source identifiers, model revisions, hashes,
unfinished-generation flags, and scoring and sensitivity rules.

\end{document}